\documentclass[11pt,a4paper]{article}
\usepackage[hyperref]{eacl2021}
\usepackage{times}
\usepackage{latexsym}
\usepackage{graphicx}
\usepackage{enumitem}
\usepackage{todonotes}

\usepackage{microtype}

\aclfinalcopy 

\title{Are pre-trained text representations useful for multilingual and multi-dimensional language proficiency modeling?}

\author{Taraka Rama \\
 \small University of North Texas, USA \\
  \small \texttt{Taraka.Kasicheyanula@unt.edu} \\\And
  Sowmya Vajjala \\
  \small National Research Council, Canada \\
  \small \texttt{sowmya.vajjala@nrc-cnrc.gc.ca} \\}

\date{}

\begin{document}
\maketitle
\begin{abstract}
Development of language proficiency models for non-native learners has been an active area of interest in NLP research for the past few years. Although language proficiency is multidimensional in nature, existing research typically considers a single ``overall proficiency'' while building models. Further, existing approaches also considers only one language at a time. This paper describes our experiments and observations about the role of pre-trained and fine-tuned multilingual embeddings in performing multi-dimensional, multilingual language proficiency classification. We report experiments with three languages -- German, Italian, and Czech -- and model seven dimensions of proficiency ranging from vocabulary control to sociolinguistic appropriateness. Our results indicate that while fine-tuned embeddings are useful for multilingual proficiency modeling, none of the features achieve consistently best performance for all dimensions of language proficiency\footnote{All code, data and related supplementary material can be found at: \url{https://github.com/nishkalavallabhi/MultidimCEFRScoring}}.
\end{abstract}

\section{Introduction}
\label{sec:intro}
Automated Essay Scoring (AES) is the task of grading test taker writing using computer programs. It has been an active area of research in NLP for the past 15 years. Although most of the existing research focused on English, recent years saw the development of AES models for second language proficiency assessment for non-English languages, typically modeled using the  Common European Framework of Reference (CEFR) reference scale \cite{CEFR-02} in Europe.

Most of the past research focused on monolingual AES models. However, the notion of language proficiency is not limited to any one language. As a matter of fact, CEFR \cite{CEFR-02} provides language agnostic guidelines to describe different levels of language proficiency, from A1 (beginner) to C2 (advanced). Hence, a universal, multilingual language proficiency model is an interesting possibility to explore. From an application perspective, it will be useful to know if one can achieve cross-lingual transfer and build an AES system for a new language without or with little training data. \newcite{Vajjala.Rama-18} explored these ideas with basic features such as n-grams and POS tag ratios. The usefulness of large, pre-trained multilingual models (with or without fine-tuning) from recent NLP research has not been studied for this task, especially for non-English languages.  

Further, AES research generally considers language proficiency as a single construct. However, proficiency encompasses multiple dimensions such as vocabulary richness, grammatical accuracy, coherence/cohesion, usage of idioms etc. \cite{Attali.Bustein-04}. CEFR guidelines also provide language proficiency rubrics for individual dimensions along with overall proficiency for A1--C2. Modeling multiple dimensions instead of a single ``overall proficiency'' could result in a more fine-grained assessment for offering specific feedback.

Given this background, we explore the usefulness of multilingual pre-trained embeddings for training multi-dimensional language proficiency scoring models for three languages -- German, Czech and Italian. 
The main contributions of our paper are listed below:
\begin{itemize}
    \item We address the problem of multi-dimensional modeling of language proficiency for three (non-English) languages.
    \item We explore whether large pre-trained, multilingual embeddings are useful as feature representations for this task with and without fine-tuning. 
    \item We investigate the possibility of a universal multilingual language proficiency model and zero-shot cross lingual transfer using embedding representations. 

\end{itemize}
The paper is organized as follows. Section~\ref{sec:relw} briefly surveys the related work. Section~\ref{sec:methods} describes our corpus, features, and experimental settings. Section~\ref{sec:expts} discusses our results in detail. Section~\ref{sec:summary} concludes the paper with pointers to future work. 

\section{Related Work}
\label{sec:relw}
Automated Essay Scoring (AES) is a well-researched problem in NLP and has been applied to real-world language assessment scenarios for English \cite{Attali.Bustein-04}. A wide range of  features such as document length, lexical/syntactic n-grams, and features capturing linguistic aspects such as vocabulary, syntax and discourse are commonly used \cite{Klebanov.Flor-13,Phandi.Chai.ea-15,Zesch.Wojatzki.ea-15}. In the recent past, different forms of text embeddings and pre-trained language models have also been explored \cite{Alikaniotis.Yannakoudakis.ea-16,Dong.Zhang-16,Mayfield.Black-20} along with approaches to combine linguistic features with neural networks \cite{Shin-18,Liu.Xu.ea-19}. \newcite{Ke.Ng-19} and \newcite{Klebanov.Madnani-20} present the most recent surveys on the state of the art in AES (focusing on English).

In terms of modeling, AES has been modeled as a classification, regression, and ranking problem, with approaches ranging from linear regression to deep learning models. Some of the recent work explored the usefulness of multi-task learning \cite{Cummins.Rei-18,Berggren.Rama.ea-19} and transfer learning \cite{Jin.He.ea-18,Ballier.Canu.ea-20}. Going beyond approaches that work for a single language, \newcite{Vajjala.Rama-18} reported on developing methods for multi- and cross-lingual AES. 

Much of the existing AES research has been focused on English, but there is a growing body of research on other European languages: German \cite{Hancke.Meurers-13}, Estonian \cite{Vajjala.Loo-14}, Swedish \cite{Pilan.Alfter.ea-16}, Norwegian \cite{Berggren.Rama.ea-19} which explored both language specific (e.g., case markers in Estonian) as well as language agnostic (e.g., POS n-grams) features \cite{Vajjala.Rama-18} for this task. However, to our knowledge, the use of large pre-trained language models such as BERT \cite{Devlin.Chang.ea-18} has not been explored yet for AES in non-English languages. 

Further, most of the approaches focused on modeling language proficiency as a single variable. Although there is some research focusing on multiple dimensions of language proficiency \cite{Lee.Gentile.ea-09,Attali.Sinharay-15,Agejev.Snajder-17,Mathias.Bhattacharyya-20}, none of them focused on non-English languages or used recent multilingual pre-trained models such as BERT. In this paper, we focus on this problem of multi-dimensional modeling of language proficiency for three languages---German, Italian, and Czech---and explore whether recent research on multilingual embeddings can be useful for non-English AES. 

\section{Datasets and Methods}
\label{sec:methods}
In this section, we describe the corpus, features, models, and implementation details. We modeled the task as a classification problem and trained individual models for each of the 7 dimensions of language proficiency. The rest of this section describes the different steps involved in our approach in detail. 
\subsection{Corpus}
\label{subsec:corpus}
In this paper, we employed the publicly available MERLIN corpus \cite{Boyd.Hana.ea-14}, which was also used in the experiments reported in some past research \cite{Hancke-13,Vajjala.Rama-18} and in the recently conducted REPROLANG challenge \cite{Branco.Calzolari.ea-20}. The MERLIN  corpus\footnote{Available here for download: \url{https://merlin-platform.eu/C_download.php}} contains CEFR scale based language proficiency annotations for texts produced by non-native learners in three languages: German, Czech, and Italian in seven dimensions which are described below: 

\begin{enumerate}
    \item \textbf{Overall proficiency} is the generic label expected to summarize the language proficiency across different dimensions. 
    \item \textbf{Grammatical accuracy} refers to the usage and control over the language's grammar.
    \item \textbf{Orthographic control} refers to the aspects of language connected with writing such as punctuation, spelling mistakes etc. 
    \item \textbf{Vocabulary range} refers to the breadth of vocabulary use, including phrases, idiomatic expressions, colloquialisms etc. 
    \item \textbf{Vocabulary control} refers to the correct and appropriate use of vocabulary. 
    \item \textbf{Coherence and Cohesion} refers to the ability to connect different parts of the text using appropriate vocabulary (e.g., connecting words) and creating a smoothly flowing text. 
    \item \textbf{Sociolinguistic appropriateness} refers to the awareness of language use in different social contexts. For example, using proper form of introduction, ability to express both in formal as well as informal language, understanding the sociocultural aspects of language use etc. 
\end{enumerate}

Detailed description of a dimension at each CEFR level is provided in the structured overview of CEFR scales document \cite{CEFR-02}. In the MERLIN corpus, these annotations were prepared by human graders who were trained on these well defined rubrics. More details on the examination setup, grade annotation guidelines, rating procedure, inter-rater reliability and reliability of rating measures can be found in the project documentation \cite{Barenfanger-13}. We used the texts and their universal dependency parsed versions -- shared by \newcite{Vajjala.Rama-18} -- consisting of 2266 documents in total (1029 German, 803 Italian, 434 for Czech). 

The German corpus had A1--C1, Italian corpus had A2--B1, and Czech had A2--B2 levels for the overall proficiency category. More CEFR levels were represented in the corpus for other proficiency dimensions.\footnote{More details on the corpus distribution can be found in MERLIN documentation, and the result files we shared as supplementary material contain CEFR level distributions for all the classification scenarios, for all languages.} In this paper, we treat the annotated labels in the corpus as the gold standard labels.

\paragraph{Missing labels} Less than 10 documents had an annotation of $1$ instead of the CEFR scale (A1--C2) for some of the dimensions. The documentation did not provide any reason behind this label assignment and we removed them from our experiments. In the case of German and Italian, for less than ten documents, some individual dimensions had a score of ``0'' while the overall rating was A1. For these documents, we treated ``0'' score as A1 rating for that dimension. In the case of Czech, about half of the documents ($247$) for the sociolinguistic appropriateness dimension had a score of ``0''. The corpus manual does not provide any explanation for the missing annotation, therefore, we excluded this dimension from all experiments involving Czech data.

\paragraph{Inter-dimensional correlations} \newcite{Barenfanger-13}'s analyses on MERLIN corpus show that correlations among the different dimensions (including overall proficiency) range from $0.2-0.8$ in different languages.\footnote{For details: Refer to Table 4 for Czech, Table 11 for German, Table 17 for Italian in \newcite{Barenfanger-13}.} In general, correlations between any two dimensions and specifically with overall proficiency dimension are higher for German and Italian than for Czech. There is no consistent high correlation of overall proficiency with any single dimension. The variations show that these individual dimensions as are indeed different from each other as well as overall proficiency dimension, and we could expect that a model trained on one dimension need not necessarily reflect on the language proficiency of the test taker in another dimension. This further motivates our decision to explore a multi dimensional proficiency assessment approach in this paper. 

\subsection{Features}
\label{subsec:features}
One of the goals of the paper is to examine if text representations computed from large, pre-trained, multilingual models such as mBERT \cite{Devlin.Chang.ea-18} and LASER \cite{artetxe2019massively} are useful for the AES task. We trained classifiers based on these two pre-trained models and compare them with two previously used features---document length baseline and the n-gram features used in \newcite{Vajjala.Rama-18}. All the features are described below:

\begin{itemize}
\item \textbf{Baseline}: Document length (number of tokens) is a standard feature in all AES approaches \cite{Attali.Bustein-04}.
\item \textbf{Lexical and syntactic features}: n-grams ($1\le n \le 5$) of Word, Universal POS tags (UPOS) from the Universal Dependencies project \cite{Nivre.deMarneffe.ea-16}, dependency triplets consisting of the head POS label, dependent POS label, and the dependency label extracted using UDPipe \cite{Straka.Hajic.ea-16}.
While word n-grams are useful only for monolingual setting, the syntactic level n-grams were used in multi/cross-lingual scenarios as well, as they are all derived from the same tagset/dependency relations. 
\item \textbf{LASER embeddings} map a sentence in a source language to a fixed dimension ($1024$) vector in a common cross-lingual space allowing us to map the vectors from different languages into a single space. Since the number of sentences in an essay is variable, we map each sentence in the segmented text to a vector and then compute the average of the vectors to yield a $1024$ dimension representation as our feature vector.
\item \textbf{mBERT}: We apply the 12-layer pre-trained multilingual BERT (trained on Wikipedias of 104 languages with shared word-piece vocabulary) for mapping an essay (truncated to $400$ tokens which is the upper bound of the length for 93\% of the documents) into a $768$ dimension vector. Specifically, we use the vector for the \texttt{CLS} token from the final layer as the feature vector for non-finetuned classification experiments. We used the MERLIN corpus texts to do task specific fine-tuning of mBERT. 
\end{itemize} 

It is possible to use other representations such as using average of the tokens' embeddings of the last layer instead of using CLS token for mBERT, or explore other recent pre-trained mono-/multilingual representations. Our goal is not to explore the best representation but rather test if a representative approach could be used for this problem. To our knowledge, only \newcite{Mayfield.Black-20} studied the application of BERT for AES in English, and its utility in the context of non-English and multilingual AES models has not been explored. 

Although, it is possible to use the ``domain'' features such as spelling/grammar errors which are commonly seen in AES systems, our goal in this paper is to explore how far we can go with the representations without any language specific resources for this task. Considering that such representations are expected to capture different aspects of language \cite{Jawahar.Sagot.ea-19,Edmiston-20}, we could hypothesize that some of the domain specific features are already captured by them. 

\subsection{Models and Evaluation}
\label{subsec:models}
As discussed in Section~\ref{sec:intro}, our motivation in this paper is to evaluate whether pre-trained multilingual embedding representations are useful for performing multidimensional AES, whether they can be used to achieve a universal representation for this task (multilingual), as well as transfer from one language to another (cross-lingual) and if the pre-trained embedding representations can be transferred to the AES task (\textit{fine-tuning}). To explore this, we trained mono/multi/cross lingual classification models using each of the features described in Section \ref{subsec:features}, for each of the 7 dimensions. 

All the traditional classification models based on n-grams, LASER and mBERT were tested using traditional classification algorithms: Logistic Regression, Random Forests, and Linear SVM. The mBERT fine-tuned model consists of a softmax classification layer on top of the \texttt{CLS} token's embedding. We used the MERLIN corpus texts to fine-tune mBERT for this task in all the three classification scenarios. 

We evaluate the classifiers in monolingual and multilingual scenarios through stratified five-fold validation where the distribution of the labels is preserved across the folds. Owing to the nature of the corpus and the presence of unbalanced classes in all the languages and dimensions in the dataset, we used weighted F$_1$ score to compare model performance, as was done in the REPROLANG challenge \cite{Branco.Calzolari.ea-20}. In the cross-lingual scenario, we trained on the German dataset and tested separately on Czech and Italian languages. 

\subsection{Implementation}
All the POS and dependency n-gram features were computed using UDPipe \cite{Straka.Hajic.ea-16}. All the traditional classification models were implemented using the Python library \texttt{scikit-learn} \cite{Pedregosa.Varoquax.ea-11}  with the default settings. LASER embeddings were extracted using the python package \texttt{laserembeddings}.\footnote{\url{https://pypi.org/project/laserembeddings/}} 
The extraction of mBERT embeddings and fine-tuning was performed using the \texttt{Hugging Face} library and PyTorch.\footnote{\url{https://huggingface.co/transformers/v2.2.0/model_doc/bert.html\#bertforsequenceclassification}} The code, processed dataset, and detailed result files are uploaded as supplementary material with this paper.

\section{Experiments and Results}
\label{sec:expts}

\begin{figure*}[htb!]
    \centering
    \includegraphics[width=0.9\textwidth,height=0.55\textwidth]{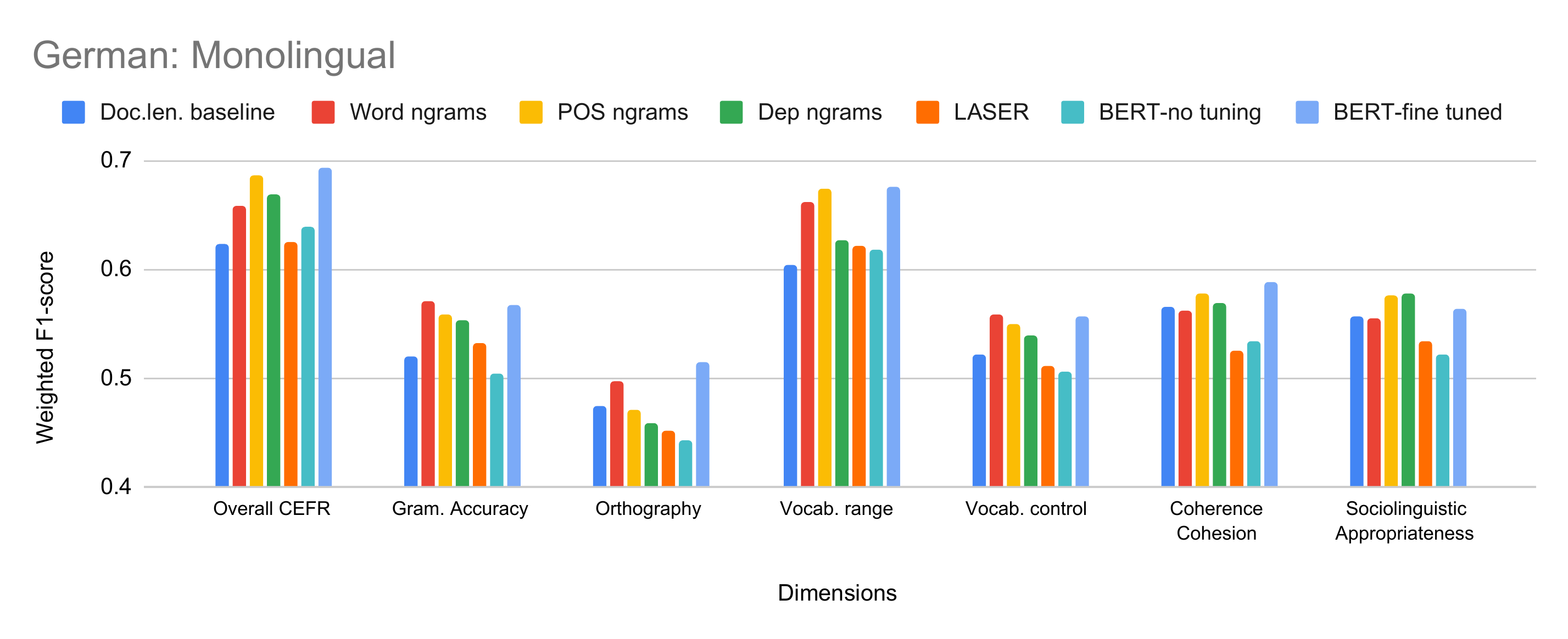}
    \caption{German monolingual five-fold validation results. All \textit{POS} and \textit{Dep} n-grams are based on Universal Dependencies framework.} 
    \label{fig:germanmono}
\end{figure*}

\begin{figure*}[htb!]
    \centering
    \includegraphics[width=0.9\textwidth,height=0.55\textwidth]{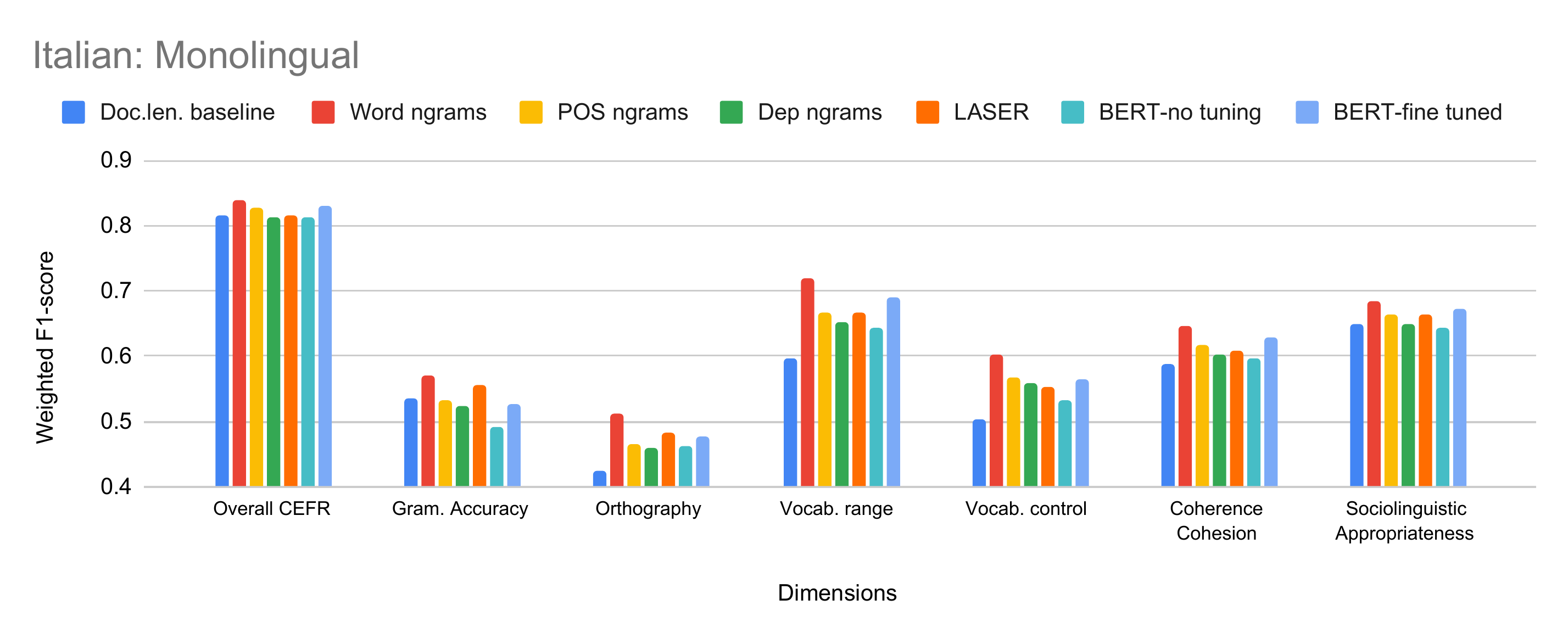}
    \caption{Italian monolingual five-fold validation results}
    \label{fig:italianmono}
\end{figure*}

\begin{figure*}[htb!]
    \centering
    \includegraphics[width=0.9\textwidth,height=0.55\textwidth]{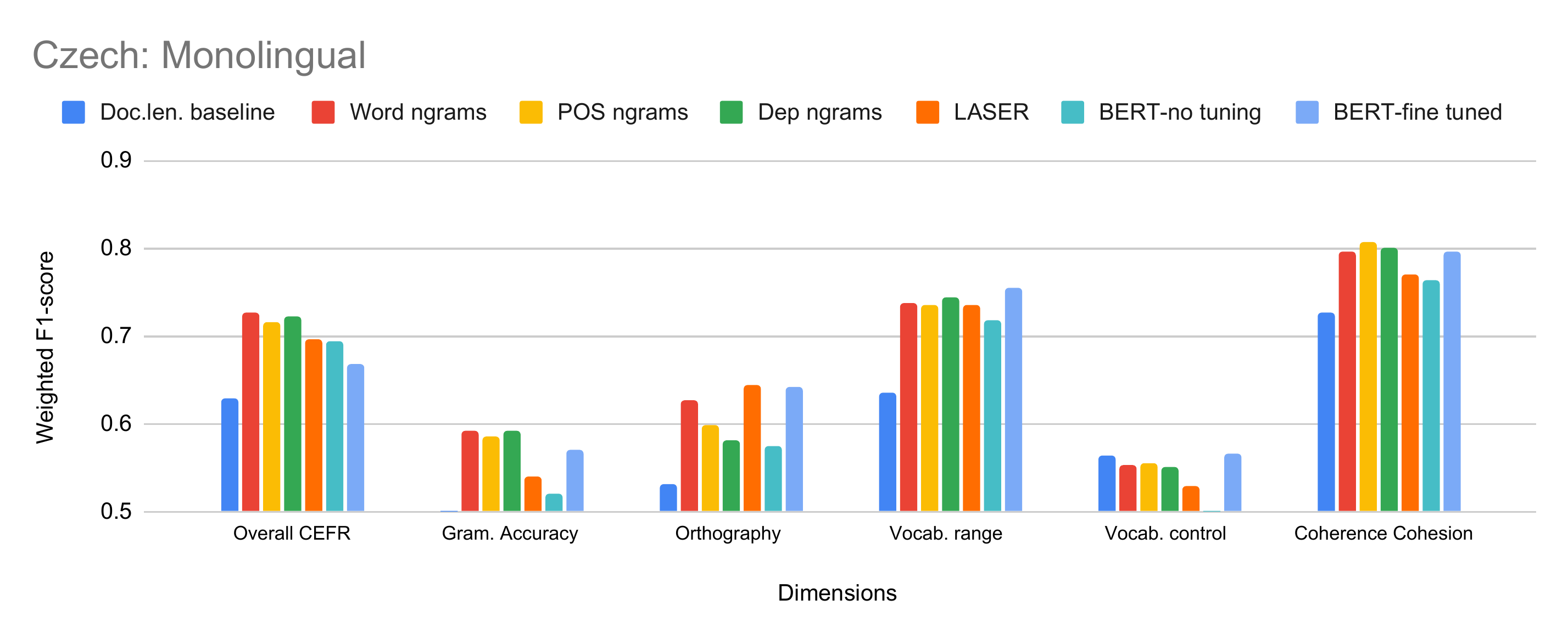}
    \caption{Czech monolingual five-fold validation results}
    \label{fig:czechmono}
\end{figure*}

As mentioned earlier, we trained monolingual, multilingual and cross-lingual classification models for all the seven proficiency dimensions. We report results with logistic regression, which performed the best in most of the cases. The results for the other classifiers such as Random Forest and Linear SVM are provided in the supplementary material.

\subsection{Monolingual Classification}
Figures \ref{fig:germanmono}, \ref{fig:italianmono} and \ref{fig:czechmono} show the results of monolingual classification for German, Italian and Czech respectively for all the feature sets and proficiency dimensions. 

\paragraph{German} The fine-tuned BERT model performs the best (from Figure~\ref{fig:germanmono}) for the Overall CEFR proficiency prediction dimension closely followed by  POS n-grams. Except for the Vocabulary Range dimension, none of the other dimensions seem to perform on par with Overall proficiency in terms of absolute numbers, though. Fine-tuned mBERT performs the best for Orthographic control dimension, where the rest of the feature sets performed rather worse. Overall, these  results seem to indicate that all our features only capture the `Overall proficiency' dimension well, and to some extent the `Vocabulary Range' dimension. All features perform rather poorly at the task of prediction of orthographic control. 

\paragraph{Italian} The word n-grams perform the best for Overall Proficiency closely followed by POS n-grams and fine-tuned mBERT model. There is not much variation among the features, with little improvement over the strong document length baseline for any feature group. Further, the performance on other dimensions seems far worse than Overall Proficiency, compared to German. Orthographic control is the worst performing dimension even for Italian. Word n-grams are the best feature representation across all dimensions for Italian. Although mBERT fine-tuning improved the performance over non fine-tuned version, both LASER and mBERT based models don't perform better than word or POS n-grams in any dimension. Thus, while there are some similarities between German and Italian classification, we also observe some differences. 

\paragraph{Czech} Across all the dimensions, the results (Figure~\ref{fig:czechmono}) for Czech are different from German and Italian. The performance of the different systems on  Coherence/Cohesion dimension is much better than the Overall Proficiency. Orthographic Control, which seemed to be the worst modeled dimension for German and Italian, does better than grammatical accuracy and vocabulary control. There is a larger difference between the baseline performance and the best performance for most of the dimensions, than it was for German and Italian. 

The main conclusions from the monolingual classification experiments are as follows: 
\begin{itemize}
\item The feature groups don't capture multiple dimensions of proficiency well and there is no single feature group that works equally well across all languages. 
\item Pre-trained and fine-tuned text representations seem to perform comparably to traditional n-gram features in several language-dimension combinations.
\end{itemize}

One possible reason for the variation across dimensions could be that the corpus consists of texts written by language learners, coming from various native language backgrounds. It is possible that there are no consistent n-gram patterns in various dimensions to capture due to this characteristic. Further, models such as LASER and mBERT are pre-trained on well formed texts, and may not be able to capture the potentially erroroneous language patterns in MERLIN texts. We can hypothesize that the overall proficiency label potentially captures some percent of each dimension, and is probably easier to model than others. However, even this hypothesis does not hold for the case of Czech, where Coherence/Cohesion dimension perhaps much better than the overall proficiency. Clearly, more analysis and experiments are needed to understand these aspects. The current set of experiments indicate that it is a worthwhile future direction to pursue.  

\subsection{Multilingual Classification}
\begin{figure*}[htb!]
    \centering
    \includegraphics[width=\textwidth,height=0.45\textwidth]{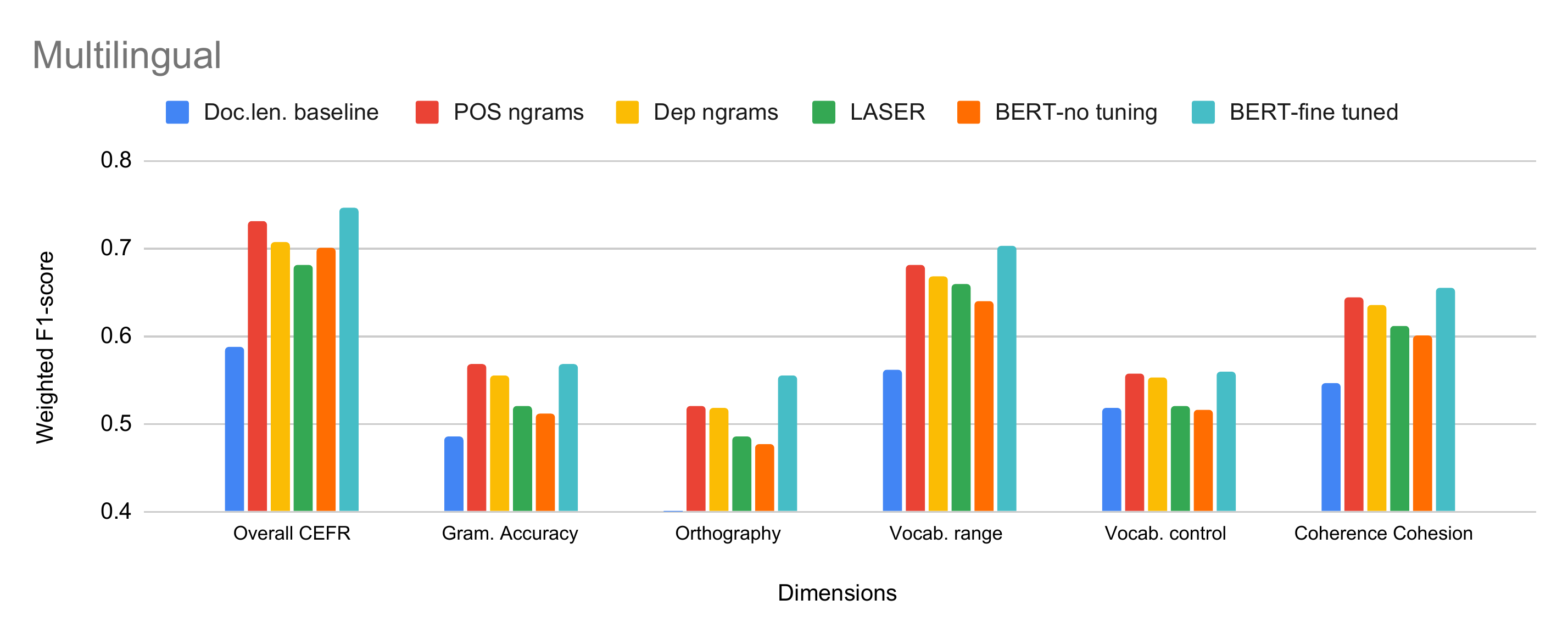}
    \caption{Multi-dimensional, Multilingual language proficiency classification. The \texttt{doclen} baseline for Orthography domain is $0.3956$ which is less than the minimum threshold of $0.4$.}
    \label{fig:multi}
\end{figure*}

In multilingual classification, we work with a single dataset formed by combining the essays from all the three languages. We trained and tested classifiers for all combinations of feature sets and dimensions on the single large dataset. Since CEFR guidelines for language proficiency are not specific to any one language, we would expect multilingual models to perform on par with individual monolingual models. The results of our multilingual experiments are given in Figure~\ref{fig:multi}.

Our results show that the fine-tuned mBERT model performs the best on most of the dimensions, closely followed by the UPOS n-grams features. To understand the relation between the multilingual model and its constituents, we looked at how each language fared in this model. For overall proficiency dimension, for example, the best result is achieved with fine-tuned classifier based on mBERT ($0.745$), which is closer to the average of the results from the three monolingual models. While German ($0.693$ in monolingual vs $0.683$ in multilingual) and Italian ($0.829$ vs $0.826$) saw a slight dip in the multilingual setup, Czech ($0.669$ vs $0.718$) saw a 5 point increase due to multilingual classification.

Clearly, multilingual classification is a beneficial setup for languages with lower monolingual performance or less data, without compromising on those languages that had better performance. However, there is still a lot of performance variation in terms of absolute numbers across dimensions. As with the case of monolingual models, we can potentially attribute this to the fact that we are dealing with a relatively smaller sized dataset in MERLIN, with texts written by people with diverse native language backgrounds, although more experiments are needed in this direction to confirm this.


\subsection{Crosslingual Classification}
Here, we train different classification models on German, and test them on Italian and Czech. We chose German for the training side since German has the largest number of essays in MERLIN corpus. In the case of mBERT, we performed fine-tuning on German part of the corpus, and tested the models on Italian and Czech texts respectively. The goal of this experiment is to test if there are any universal patterns in proficiency across the languages and understand if zero-shot cross-lingual transfer is possible for this task. 

UPOS n-grams consistently performed better than other features for most of the dimensions, in both the cross lingual setups. There is more performance variation among the different dimensions for Italian compared to Czech. In the case of Czech, similar to the monolingual case, the Coherence/cohesion dimension achieved superior performance than others, even with the baseline document length feature. This is a result worth considering further qualitative analysis in future. More details on the results of this experiment can be found in the Figures folder in the supplementary material. Our cross-lingual experiments seem to indicate that the embedding representations we chose are not useful for zero-shot learning, and that UPOS n-grams may serve as a strong baseline for building AES systems with new languages. 





\subsection{Error Analysis} 
We observed substantial performance differences across features/dimensions/languages in various experimental settings. While we don't have a procedure to understand the exact reasons for this yet, examining the confusion matrices (provided in the supplementary material) may give us some insights into the nature of some of these differences. Therefore, we manually inspected a few confusion matrices, posing ourselves three questions:
\begin{enumerate}
    \item How does a given feature set perform across different dimensions for a given language? 
    \item How do different features perform for a single dimension for a given language?
    \item How does a given feature set perform for a given dimension among the three languages? \end{enumerate}
In all these cases, we did not notice any major differences, and the confusion matrices followed the expected trend (observed in previous research) -- immediately proximal levels such as A2/B1 or A1/A2 are harder to distinguish accurately as compared to distant levels such as A1/B2 along with the expected observation that levels with larger representation have better results. It is neither possible to cover all possible combinations nor is it sufficient to gain more insights into the models just by looking at confusion matrices alone. Carefully planned interpretable analyses should be conducted in future to understand these differences further.

\section{Summary and Conclusions}
\label{sec:summary}

In this paper, we reported several experiments exploring multi-dimensional CEFR scale based language proficiency classification for three languages. Our main conclusions from these experiments can be summarized as follows:
\begin{enumerate}\itemsep0.5ex
\item UPOS n-gram features perform consistently well for all languages in monolingual classification scenarios for modeling ``overall proficiency'', closely followed by embedding features in most language-dimension combinations.
\item Fine-tuned large pre-trained models such as mBERT are useful language representations for multilingual classification, and languages with low monolingual performance benefit from a multilingual setup. 
\item UPOS features seem to provide a strong baseline for zero-shot cross lingual transfer, and fine-tuning was very not useful in this case.  
\item None of the feature groups consistently perform well across all dimensions/languages/classification setups.
\end{enumerate}

The first conclusion is similar to \cite{Mayfield.Black-20}'s conclusion on using BERT for English AES. However, these results need not be interpreted as a ``no'' to pre-trained models. Considering that they are closely behind n-grams in many cases and were slightly better than them for German, we believe they are useful to this task and more research needs to be done in this direction exploring other language models/fine-tuning options.

Pre-trained and fine-tuned models are clearly useful in a multilingual classification setup, and it would be an interesting new direction to pursue for this task. As a continuation of these experiments, one can look for a larger CEFR annotated corpus for a language such as English, and explore multilingual learning for languages with lesser data. 

The results from the experiments presented in this paper highlight the inherent difficulty in capturing multiple dimensions of language proficiency through existing methods, and the need for more future research in this direction. An important direction for future work is to develop better feature representations that capture specific dimensions of language proficiency, which can potentially work for many languages. Considering that all the dimensions share some commonalities and differences with each other, multi-task learning is another useful direction to explore. 
\bibliography{bea2021.bib}
\bibliographystyle{acl_natbib}

\appendix

\section{Supplemental Material}
\label{sec:supplemental}

The code and data for these experiments can be found at: \url{https://github.com/nishkalavallabhi/MultidimCEFRScoring}


\end{document}